\documentclass[conference]{IEEEtran}
\IEEEoverridecommandlockouts
\usepackage{cite}
\usepackage{amsmath,amssymb,amsfonts}
\usepackage{algorithmic}
\usepackage{graphicx}
\usepackage{textcomp}
\usepackage{xcolor}
\usepackage{multirow}
\usepackage{algorithm}
\usepackage{stfloats}
\usepackage{algorithmic} 
\usepackage{subcaption} 
\usepackage{amsmath}
\usepackage{bm}
\usepackage[a4paper, total={184mm,239mm}]{geometry}

\def\BibTeX{{\rm B\kern-.05em{\sc i\kern-.025em b}\kern-.08em
    T\kern-.1667em\lower.7ex\hbox{E}\kern-.125emX}}

\begin{document}

\title{Neural Network-Based CLEAN for Channel Modeling  for Unknown and Variable Number of Sources}
\title{Neural Network-Assisted CLEAN for Channel Modeling in Low-SNR Regimes}


\author{
\IEEEauthorblockN{Chaofan Deng, Linyu Sun, Jaeho Lee, Arijit Raychowdhury}
\IEEEauthorblockA{School of Electrical and Computer Engineering, Georgia Institute of Technology, Atlanta, GA, USA\\
Emails: \{cdeng63, lsun327, jlee3941\}@gatech.edu, arijit.raychowdhury@ece.gatech.edu}
}
\maketitle

\begin{abstract}
Accurate multipath parameter estimation is critical for modern wireless communication systems, particularly in challenging low-SNR environments. Traditional Maximum Likelihood Estimation algorithms, such as CLEAN, provide high-resolution parameter extraction but suffer from prohibitive computational complexity due to exhaustive grid search. Conversely, purely data-driven deep learning approaches lack physical grounding and struggle to generalize across variable multipath densities and off-grid parameters. To address these limitations, this paper proposes Neural Network-Assisted CLEAN (NN-CLEAN), a hybrid framework that embeds a multi-head residual network directly into the iterative CLEAN extraction loop. By replacing the exhaustive grid search with rapid, parallelizable forward passes while delegating residual subtraction to exact mathematical models, NN-CLEAN isolates physical multipath parameters without accumulating non-physical errors. Extensive Monte Carlo simulations demonstrate that NN-CLEAN achieves estimation accuracy exceeding $96$\% at $5$~dB SNR, matching the traditional Grid-Search CLEAN (GS-CLEAN) baseline, while providing a massive reduction in computational complexity and substantially outperforming subspace methods and standalone one-shot neural networks. Crucially, NN-CLEAN exhibits a near-flat scaling in execution runtime and memory consumption as batch sizes increase. This highly efficient parallelization establishes NN-CLEAN as a robust, real-time solution for channel estimation in MIMO systems.
\end{abstract}

\begin{IEEEkeywords}
Neural Network, Deep Learning, CLEAN, Channel Sounding, DOA Estimation, Low-SNR
\end{IEEEkeywords}

\section{Introduction}

Accurate channel sounding and Direction of Arrival (DOA) estimation are fundamental to modern wireless systems, particularly for emerging applications such as Integrated Sensing and Communication (ISAC) and advanced spatial multiplexing. While these technologies offer transformative potential, they face severe practical constraints in complex, dynamic electromagnetic environments where signals are heavily influenced by severe noise and multipath interference. Consequently, the practical utility of these systems hinges on ensuring robust parameter extraction, specifically within the low Signal-to-Noise Ratio (SNR) regime \cite{ISAC_ROBUST,ISCAS_low_snr}.

Conventional high-resolution parameter estimation relies heavily on model-driven signal processing algorithms. Subspace-based methods, such as MUSIC \cite{music, 3d_music} and ESPRIT \cite{esprit, 3d_esprit}, offer high accuracy but are fundamentally limited by the requirement for uncorrelated and non-coherent sources \cite{hussain2023low}. Furthermore, their performance degrades sharply in low-SNR environments due to the subspace swap phenomenon, where noise eigenvectors become indistinguishable from signal eigenvectors, leading to catastrophic estimation failures \cite{xu2022high}.

In contrast, Maximum Likelihood Estimation (MLE) methods, such as CLEAN \cite{seun_clean, grand_clean}, bypass these coherence restrictions and offer improved robustness against source correlation \cite{yang2023robust}. However, these algorithms typically require exhaustive grid searches across the spatial spectrum, leading to exorbitant computational overhead and memory usage that fail to scale efficiently \cite{mle_scale_issue}. Moreover, even MLE methods struggle under low-SNR conditions, as heavy noise increases the probability of the algorithm converging to local optima or dominant sidelobes rather than the true global maximum \cite{mle_low_snr}.

Recent studies have increasingly applied Deep Learning (DL) to address these computational bottlenecks by treating DOA estimation as a data-driven regression or classification task  \cite{NN1, NN2, NN4, NN5, NN_off_grid, NN_vary_MPC}. While pure DL models offer rapid inference, their reliability is inherently constrained by their training assumptions. Most existing architectures simplify the estimation problem by assuming incoming signals align perfectly with a predefined discrete spatial grid. Consequently, when deployed in continuous real-world environments, they suffer from severe performance degradation due to \textit{grid mismatch}, where off-grid signals lead to significant estimation bias and spectral leakage \cite{NN_off_grid, off_grid_iscas}. Furthermore, these one-shot architectures struggle in dynamic environments where the number of multipath components (MPCs), denoted as $N_{\text{MPC}}$, fluctuates. Because their output layers have fixed dimensions, they cannot naturally adapt to variable MPC densities  \cite{NN_vary_MPC} —a fundamental \textit{Out-of-Distribution (OOD)} failure. Without a mathematically rigorous physical model to anchor their predictions, purely data-driven approaches remain highly sensitive to noise. This lack of physical grounding frequently results in the identification of spurious signal peaks in low-SNR regimes or a complete failure to resolve weak MPCs that deviate from the training distribution.

These limitations underscore a fundamental gap: traditional mathematical estimators suffer from prohibitive computational costs and susceptibility to noise-induced local optima, while purely data-driven DL methods lack the physical robustness required for OOD scenarios, such as variable $N_{\text{MPC}}$ and continuous off-grid parameters. To bridge this gap, we propose Neural Network-Assisted CLEAN (NN-CLEAN), a hybrid framework that modernizes the traditional CLEAN algorithm by embedding a highly efficient residual network directly into its iterative extraction loop. Rather than predicting all MPCs simultaneously, NN-CLEAN utilizes a rapid, parallelizable forward pass solely to isolate the spatial parameters of the dominant path at each iteration. Crucially, it explicitly addresses the physical limitations of pure DL architectures by delegating the subsequent complex path gain calculation and residual subtraction to exact mathematical models governed by theoretical array manifolds and wave propagation physics. Because the extracted signal is evaluated and subtracted using true physical constraints, the iterative loop prevents the network from hallucinating spurious peaks or accumulating non-physical errors. By combining the pattern-recognition speed of deep learning with the mathematically grounded structure of MLE, NN-CLEAN ensures robust generalization to off-grid MPCs and variable $N_{\text{MPC}}$. This physically anchored framework provides a highly scalable solution that outperforms both traditional subspace methods and standalone one-shot neural networks.

The main contributions of this paper are summarized as follows:
\begin{enumerate}  
    \item \textbf{NN-Assisted MLE Framework for Low-SNR Robustness:} By leveraging a NN to guide the iterative parameter extraction, the proposed framework successfully bypasses the spurious sidelobe artifacts and local optima that frequently trap exhaustive grid-search algorithms. Furthermore, by delegating the residual subtraction to exact mathematical models, NN-CLEAN curtails the error propagation inherent to pure DL methods, yielding better estimation accuracy in challenging low-SNR regimes.  
    \item \textbf{High Scalability and Fast Inference:} NN-CLEAN resolves the severe computational bottlenecks inherent to multi-dimensional grid searches. By isolating spatial parameters through highly parallelizable forward passes, the framework maintains a remarkably flat computational runtime and memory footprint even as batch sizes scale. This amortized efficiency establishes NN-CLEAN as an ideal, real-time solution for demanding, large-scale deployments such as ISAC and centralized baseband processing.  
    \item \textbf{Robust Generalization to OOD and Off-Grid Scenarios:} Unlike standalone one-shot DL models whose reliability is strictly bounded by their training assumptions, NN-CLEAN explicitly generalizes to Out-of-Distribution scenarios. Because the NN operates locally within a dynamic, iterative loop, the framework seamlessly resolves an unknown and variable number of multipath components ($N_{\text{MPC}}$), maintaining high-fidelity estimation even for continuous, strictly off-grid spatial parameters.  
\end{enumerate}

\begin{figure}[htbp]
  \centering
  \includegraphics[width=1.0\columnwidth]{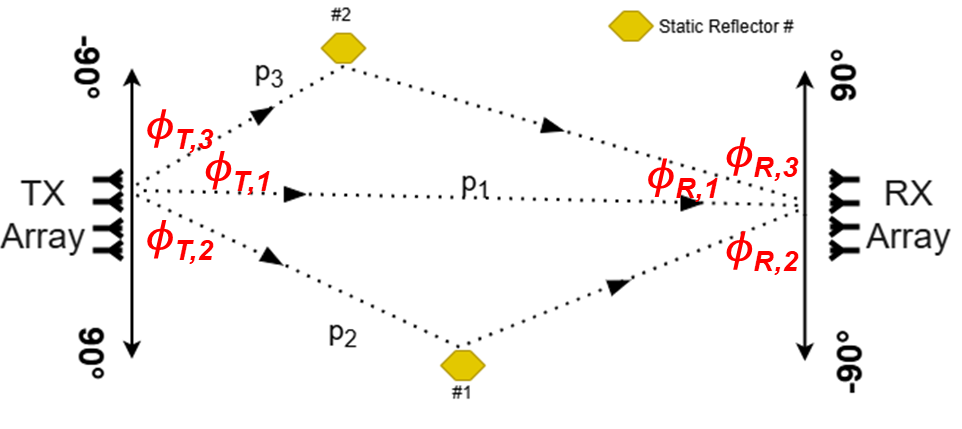}
  \caption{Channel Model Visualization}
  \label{fig:1}
\end{figure}

\section{System Description and CLEAN Algorithm}

\subsection{System Model}
Consider a quasi-static double-directional radio channel between transmitting (Tx) and receiving (Rx) Uniform Linear Arrays (ULAs), with the propagation geometry modeled as in Fig.~\ref{fig:1}. The channel comprises a line-of-sight (LOS) path and multiple scattered paths, each forming a distinct MPC. The channel transfer function matrix $\boldsymbol{H}(f_k)$, evaluated at frequency $f_k$, is modeled as the superposition of these discrete MPCs. 

Based on the framework in \cite{seun_clean}, let $b_T(\phi)$ and $b_R(\phi)$ represent the $N_T \times 1$ and $1 \times N_R$ calibrated, vertically polarized beam patterns of the transmit and receive arrays, respectively, for a given azimuth angle $\phi$. The $N_T \times N_R$ channel matrix $\boldsymbol{H}(f_k)$ is formulated as:
\begin{equation} \label{eq:channel_model}
    \boldsymbol{H}(f_k) = \sum_{l=1}^{N_{\text{MPC}}} \alpha_l b_T(\phi_{T,l}) b_R(\phi_{R,l}) \exp\left(-j 2 \pi f_k \frac{d_l}{c}\right) + \boldsymbol{N}(f_k),
\end{equation}
where $\phi_{T,l}$ and $\phi_{R,l}$ denote the Azimuth of Departure (AoD) and Azimuth of Arrival (AoA) for the $l^{\text{th}}$ MPC, respectively. The variables $d_l$ and $\alpha_l$ represent the physical travel distance and complex path gain of the $l^{\text{th}}$ path, while $c$ is the speed of light. Finally, $\boldsymbol{N}(f_k)$ denotes the additive white Gaussian noise (AWGN) matrix.

\subsection{MPC Estimation based on the CLEAN Algorithm}
Jointly estimating the parameters of all MPCs simultaneously requires a multi-dimensional estimation over a discrete parameter space, which is computationally intractable. To avoid this complexity, the CLEAN algorithm \cite{clean_algorithm} is employed as an iterative greedy matching pursuit technique that decouples the problem by extracting one dominant MPC at a time.

During the $i^{\text{th}}$ iteration, CLEAN isolates the dominant MPC through a discrete search over a three-dimensional parameter space $\Theta = \Theta_{\text{AoA}} \times \Theta_{\text{AoD}} \times \Theta_{\text{dist}}$. This finite grid is constructed by quantizing the azimuth dimensions into $N_a$ angular bins and the propagation distance into $N_d$ distance bins. The optimal parameter tuple $\{\hat{d}_i, \hat{\phi}_{T,i}, \hat{\phi}_{R,i}\} \in \hat\Theta$ is then extracted by maximizing the squared spatial-distance correlation:
\begin{equation} \label{eq:ml_estimate}
\begin{aligned}
    \{\hat{d}_i, \hat{\phi}_{T,i}, \hat{\phi}_{R,i}\} &= \arg \max_{d, \phi_T, \phi_R} \left| \sum_{k=1}^{N_f} \exp\left(-j 2 \pi f_k \frac{d}{c}\right) \right. \\
    &\quad \times \left. b_R(\phi_R) \left(\boldsymbol{H}_{\text{rem}}^{(i)}(f_k)\right)^\dagger b_T(\phi_T) \right|^2,
\end{aligned}
\end{equation}
where $\boldsymbol{H}_{\text{rem}}^{(i)}(f_k)$ denotes the residual channel transfer matrix at iteration $i$, initialized as $\boldsymbol{H}_{\text{rem}}^{(1)}(f_k) = \boldsymbol{H}(f_k)$.

Once identified, the complex path gain $\hat{\alpha}_i$ of this dominant MPC is computed via a least squares projection onto the residual channel:
\begin{equation} \label{eq:path_gain}
    \hat{\alpha}_i = \frac{\sum_{k=1}^{N_f} \exp\left(j 2 \pi f_k \frac{\hat{d}_i}{c}\right) b_T^\dagger(\hat{\phi}_{T,i}) \boldsymbol{H}_{\text{rem}}^{(i)}(f_k) b_R^\dagger(\hat{\phi}_{R,i})}{N_f \left\| b_T(\hat{\phi}_{T,i}) \right\|^2 \left\| b_R(\hat{\phi}_{R,i}) \right\|^2}.
\end{equation}

The contribution of this dominant MPC is then synthesized and subtracted from the current residual transfer function to form the updated residual for the next iteration:
\begin{equation} \label{eq:residual}
\begin{aligned}
    \boldsymbol{H}_{\text{rem}}^{(i+1)}(f_k) &= \boldsymbol{H}_{\text{rem}}^{(i)}(f_k) \\
    &\quad - \hat{\alpha}_i b_T(\hat{\phi}_{T,i}) b_R(\hat{\phi}_{R,i}) \exp\left(-j 2 \pi f_k \frac{\hat{d}_i}{c}\right).
\end{aligned}
\end{equation}

This iterative extraction and subtraction process repeats until the absolute path gain of the newly extracted MPC falls below a pre-determined energy threshold or a maximum number of paths is reached.

\textbf{Algorithm Bottlenecks:} 
The computational efficacy of traditional Grid-Search CLEAN (GS-CLEAN) is fundamentally constrained by its reliance on exhaustive parameter evaluation. For each candidate grid point, evaluating the spatial-distance correlation in (\ref{eq:ml_estimate}) requires $\mathcal{O}(N_f N_T N_R)$ operations. Quantizing our three-parameter search space ($d, \phi_T, \phi_R$) into $G$ bins per dimension inflates the per-iteration complexity to $\mathcal{O}(G^3 N_f N_T N_R)$, while incurring substantial memory overhead for precomputed array manifold dictionaries. This burden is further compounded by the curse of dimensionality; extending the framework to fully 3D spatial environments (incorporating elevation angles) or high-mobility scenarios (including Doppler shifts) expands the search domain to five or more dimensions, driving complexity to an intractable $\mathcal{O}(G^5 N_f N_T N_R)$. This cubic-to-quintic scaling renders exhaustive grid-search paradigms fundamentally impractical for real-time, low-latency applications, thereby necessitating a shift toward efficient, learning-based estimation frameworks.

\section{Proposed NN-CLEAN Algorithm}
\begin{figure}[htbp]
  \centering
  \includegraphics[width=1.0\columnwidth]{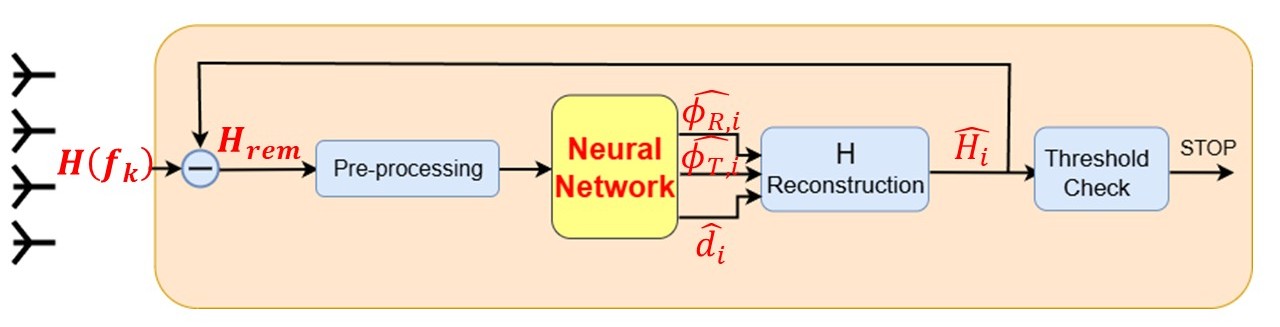}
  \caption{NN-assisted CLEAN System Overview}
  \label{fig:2}
\end{figure}

To overcome the severe computational bottlenecks of multi-dimensional grid searches, we propose NN-CLEAN. As illustrated in Fig.~\ref{fig:2}, this hybrid framework embeds a NN directly into the CLEAN iterative extraction loop, replacing exhaustive parameter evaluations with a rapid, direct inference. During the $i^{\text{th}}$ iteration, the complex residual channel matrix $\boldsymbol{H}_{\text{rem}}^{(i)}(f_k)$ is decoupled into its real and imaginary components to form the input tensor. A single forward pass maps this observation directly to the dominant MPC, yielding the parameter tuple $\hat{\Theta}_i = \{\hat{d}_i, \hat{\phi}_{T,i}, \hat{\phi}_{R,i}\}$. This streamlined approach maintains high-fidelity extraction while incurring only a fraction of the traditional memory and computational overhead.

\begin{figure}[htbp]
  \centering
  \includegraphics[width=1.0\columnwidth]{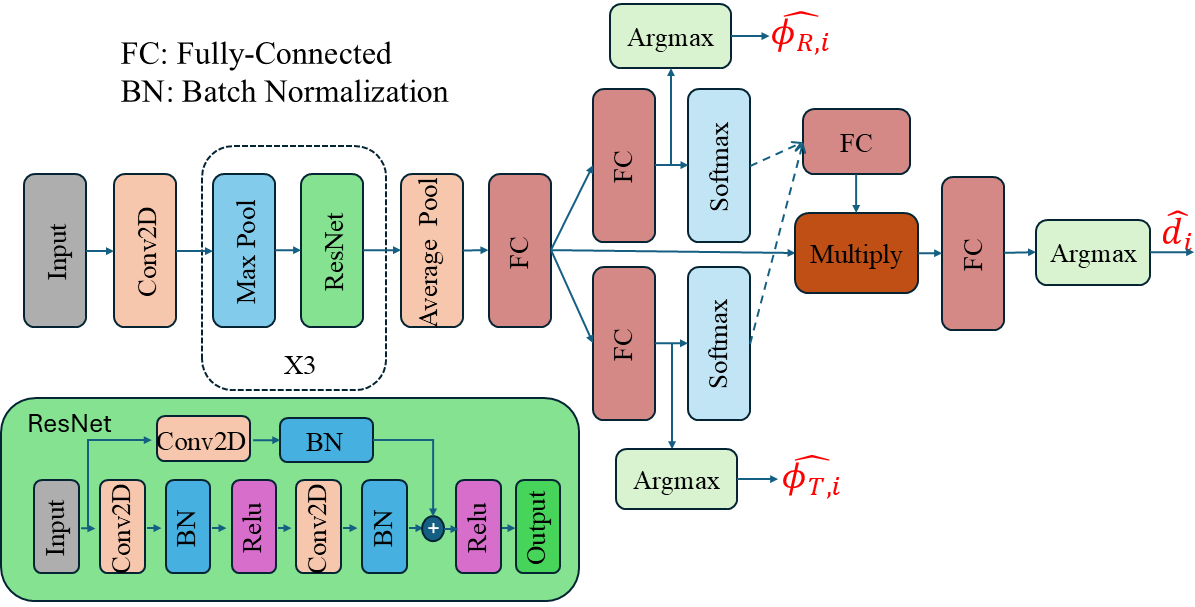}
  \caption{NN Structure Overview}
  \label{fig:3}
\end{figure}

As depicted in Fig.~\ref{fig:3}, the NN is structured as a multi-head residual classification architecture that evaluates the channel against pre-defined parameter grids. A cascaded backbone of MaxPooling and Residual Network (ResNet) efficiently compresses the frequency domain while preserving critical spatial-frequency features. The extracted latent features are then routed to independent fully connected (FC) heads, which output raw logits representing the spatial probability distributions for the AoA and AoD. Because the physical distance of a multipath component is inherently coupled with its spatial trajectory, an FC gating mechanism explicitly conditions the final distance head on these resolved angular probabilities to enforce structural consistency. Finally, an $\arg\max$ operation is applied to the output logits to select the discrete grid indices with the highest probabilities, mapping them directly to the estimated physical parameters of $\hat{\Theta}_i$.

\section{Simulation Results}
\subsection{Simulation Setup}
\subsubsection{Channel Model Setup}
The proposed system is implemented and evaluated with a synthesized $4 \times 4$ MIMO communication system operating across $N_f = 512$ uniformly spaced subcarriers in the $2$--$3$ GHz range. 

Both the TX and RX employ ULAs with half-wavelength antenna spacing evaluated at a center frequency of $f_c = 2.5$ GHz. To isolate algorithmic performance from hardware imperfections while aligning with the co-polarized system model, we assume ideal vertically polarized omnidirectional elements (e.g., ideal dipoles) with negligible mutual coupling. Consequently, the array's amplitude response is uniform across the azimuth plane, and its spatial response is governed entirely by frequency-dependent geometric phase differences. For an $N$-element array, the theoretical far-field steering vector at spatial angle $\phi$ and subcarrier $f_k$ is defined as:
\begin{equation} \label{eq:steering_vector}b(\phi, f_k) = \frac{1}{\sqrt{N}} \left[ 1, e^{-j \pi \frac{f_k}{f_c} \sin(\phi)}, \dots, e^{-j \pi (N-1) \frac{f_k}{f_c} \sin(\phi)} \right]^T .\end{equation}

For the grid-based evaluations and the generation of synthetic data, the parameter space $\Theta$ is discretized as follows:
\begin{itemize}
    \item \textbf{AoD ($\phi_T$) and AoA ($\phi_R$):} Ranging from $-90^\circ$ to $90^\circ$ with a $1^\circ$ resolution ($N_a = 181$ bins).
    \item \textbf{Distance ($d$):} Ranging from $5$~m to $15$~m with a $0.05$~m resolution ($N_d = 201$ bins).
\end{itemize}

\subsubsection{Training Setup}
The training procedure for the proposed NN-assisted CLEAN framework is detailed in Algorithm~\ref{alg:model-training}. The network is optimized using the Adam optimizer (learning rate $10^{-3}$, batch size $128$). To ensure robust generalization, training samples are dynamically synthesized on the fly. In each training step, a channel comprising $N_{\text{MPC}}=4$ MPCs is generated, with ground-truth parameters ($\phi_R$, $\phi_T$, $d$) drawn strictly from their respective \textbf{on-grid} discrete sets. The complex path gains undergo simulated Rayleigh fading ($\alpha \sim \mathcal{CN}(0, 2)$), and additive white Gaussian noise (AWGN) is injected at an $\text{SNR} \in [-5, 10]$~dB.

To capture the topological correlation between adjacent discrete bins, we apply Gaussian Label Smoothing (GLS) \cite{GLS} to the ground-truth labels. For a true on-grid parameter index $\mu$, the smoothed target probability $t_j$ at grid index $j$ is formulated as a normalized Gaussian distribution:
\begin{equation} \label{eq:gls}
    t_j = \frac{\exp\left(-\frac{1}{2} \left(\frac{j - \mu}{\sigma}\right)^2\right)}{\sum_k \exp\left(-\frac{1}{2} \left(\frac{k - \mu}{\sigma}\right)^2\right)} ,
\end{equation}
where the spread parameter is empirically set to $\sigma = 2.0$. 

Because the network extracts multipath components sequentially, a distinct loss is computed at each extraction iteration $i$. The iteration loss is defined as the weighted Categorical Cross-Entropy (CCE) between the network's logits ($\hat{\boldsymbol{y}}^{(i)}$) and the GLS targets ($\boldsymbol{t}^{(i)}$):
\begin{equation} \label{eq:step_loss}
    \mathcal{L}_{\text{iter}}^{(i)} = \text{CCE}\left(\boldsymbol{t}_{\phi_R}^{(i)}, \hat{\boldsymbol{y}}_{\phi_R}^{(i)}\right) + \text{CCE}\left(\boldsymbol{t}_{\phi_T}^{(i)}, \hat{\boldsymbol{y}}_{\phi_T}^{(i)}\right) + 3 \cdot \text{CCE}\left(\boldsymbol{t}_{d}^{(i)}, \hat{\boldsymbol{y}}_{d}^{(i)}\right) ,
\end{equation}
A weight of $3$ is applied exclusively to the distance loss. Because precise distance estimation is critical for achieving accurate phase alignment during residual cancellation, this penalty ensures the network prioritizes resolving the distance dimension.

\begin{algorithm}[htbp]
\caption{Training Procedures for NN-assisted CLEAN}
\label{alg:model-training}
\begin{algorithmic}[1]
\STATE \textbf{Initialize:} Model weights, Adam optimizer ($\text{lr} = 10^{-3}$)
\FOR{each training step $t = 1$ \textbf{to} $1,000,000$}
    \STATE Sample $\text{SNR} \sim \mathcal{U}(-5, 10)$~dB
    \STATE Generate $N_{\text{MPC}} = 4$ MPCs: $\{\phi_{R,l}, \phi_{T,l}, d_l\}_{l=1}^{N_{\text{MPC}}} \in \Theta$
    \STATE Sample complex path gains $\alpha \sim \mathcal{CN}(0, 2)$
    \STATE Synthesize channel $\boldsymbol{H}$ and add AWGN
    \STATE Generate GLS targets $\boldsymbol{t}_{\phi_R}^{(i)}, \boldsymbol{t}_{\phi_T}^{(i)}, \boldsymbol{t}_d^{(i)}$ using \eqref{eq:gls}
    
    \IF{$t \le 500,000$}
    \STATE \textit{// Phase 1 (0--50\%): Dominant path detection}
    \STATE Extract 1st path only; compute $\mathcal{L}_{\text{final}} = \mathcal{L}_{\text{iter}}^{(1)}$
\ELSIF{$t \le 750,000$}
    \STATE \textit{// Phase 2 (50--75\%): Teacher forcing}
    \STATE Extract $N_{\text{MPC}}$ paths iteratively using \textbf{ground-truth} subtraction; 
    \STATE Compute $\mathcal{L}_{\text{final}} = \frac{1}{N_{\text{MPC}}}\sum_{i=1}^{N_{\text{MPC}}} \mathcal{L}_{\text{iter}}^{(i)}$
\ELSE
    \STATE \textit{// Phase 3 (75--100\%): Autoregressive inference}
    \STATE Extract $N_{\text{MPC}}$ paths iteratively using \textbf{predicted} subtraction; 
    \STATE Compute $\mathcal{L}_{\text{final}} = \frac{1}{N_{\text{MPC}}}\sum_{i=1}^{N_{\text{MPC}}} \mathcal{L}_{\text{iter}}^{(i)}$
\ENDIF
    
    \STATE Backpropagate $\mathcal{L}_{\text{final}}$ and update model weights
\ENDFOR
\end{algorithmic}
\end{algorithm}

Finally, to mitigate error propagation inherent in iterative subtraction, the network is trained over $1,000,000$ steps using a three-phase curriculum. The final backpropagated loss, $\mathcal{L}_{\text{final}}$, adapts dynamically:
\begin{itemize}
    \item \textbf{Phase 1 (0\%--50\%):} Focuses solely on dominant path detection ($\mathcal{L}_{\text{final}} = \mathcal{L}_{\text{iter}}^{(1)}$), learning fundamental feature representations without subtraction noise.
    \item \textbf{Phase 2 (50\%--75\%):} Extracts all $N_{\text{MPC}}$ paths iteratively. Perfect residual subtraction (teacher forcing) is employed using ground-truth parameters to compute the subtracted interference, yielding an averaged loss $\mathcal{L}_{\text{final}} = \frac{1}{N_{\text{MPC}}}\sum_{i=1}^{N_{\text{MPC}}} \mathcal{L}_{\text{iter}}^{(i)}$.
    \item \textbf{Phase 3 (75\%--100\%):} Full autoregressive inference. The network subtracts its own predicted components to learn resilience against accumulated errors, utilizing the same averaged loss $\mathcal{L}_{\text{final}} = \frac{1}{N_{\text{MPC}}}\sum_{i=1}^{N_{\text{MPC}}} \mathcal{L}_{\text{iter}}^{(i)}$.
\end{itemize}

\begin{figure}[htbp]
  \centering
  \includegraphics[width=\columnwidth]{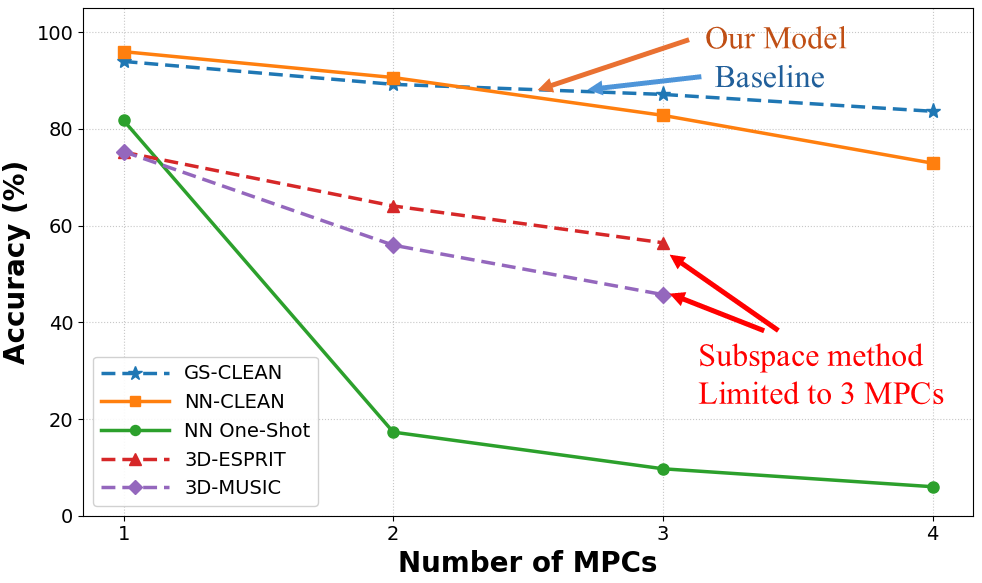}
  \caption{Estimation Accuracy Comparison Between Different Methods}
  \label{fig:4}
\end{figure}

\begin{figure*}[t!]
  \centering

  \begin{subfigure}[b]{0.32\linewidth}
    \centering
    \includegraphics[width=\linewidth]{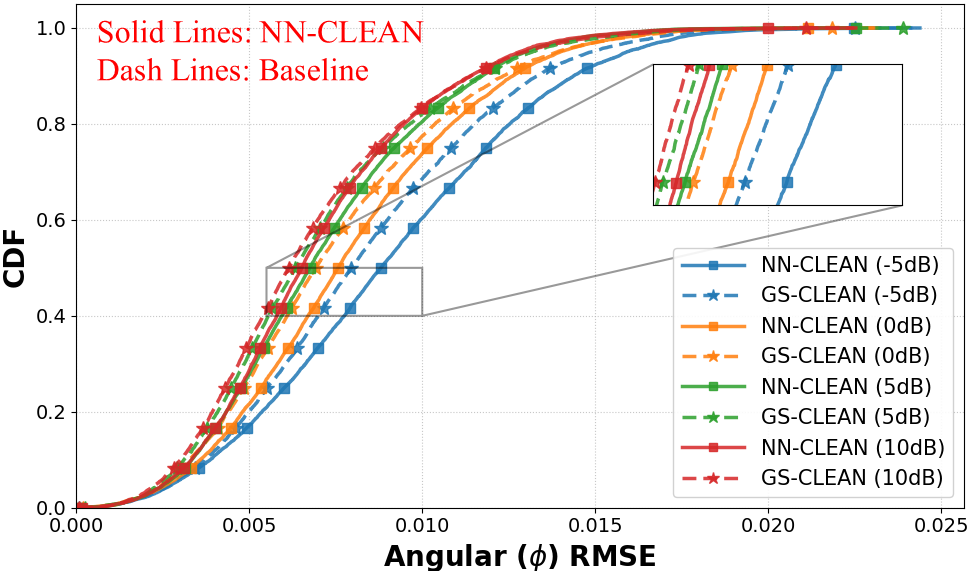}
    \caption{Spatial MCD $\eta_{\text{ang}}$}
    \label{fig:5a}
  \end{subfigure}
  \hfill 
  \begin{subfigure}[b]{0.32\linewidth}
    \centering
    \includegraphics[width=\linewidth]{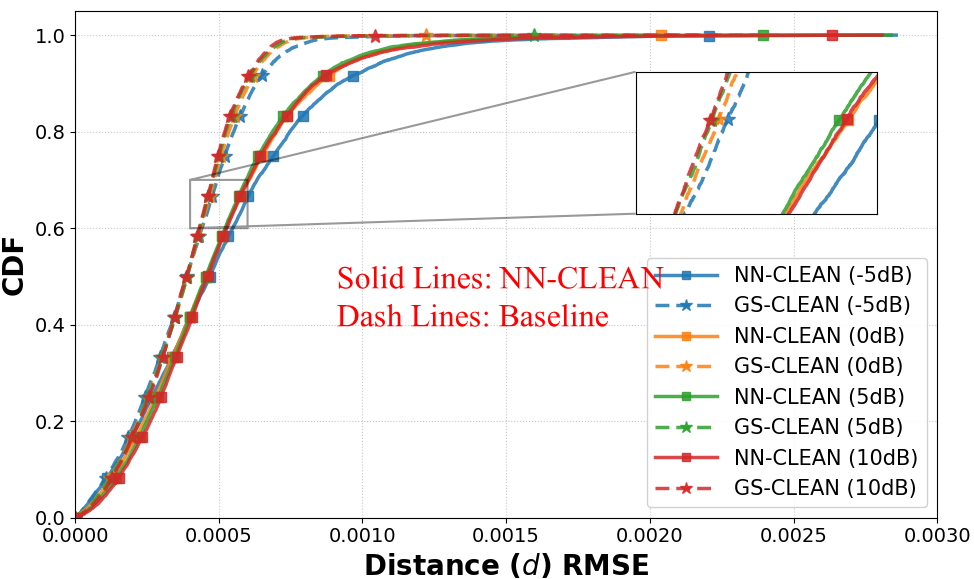}
    \caption{Distance MCD $\eta_{\text{d}}$}
    \label{fig:5b}
  \end{subfigure}
  \hfill
  \begin{subfigure}[b]{0.32\linewidth}
    \centering
    \includegraphics[width=\linewidth]{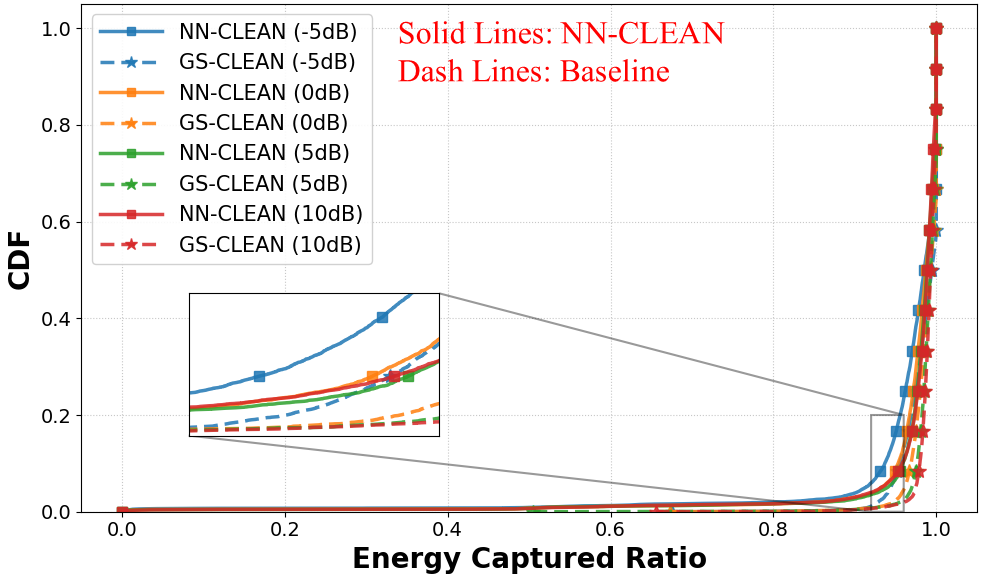}
    \caption{Energy Captured Ratio}
    \label{fig:5c}
  \end{subfigure}

  \caption{CDF comparisons for spatial MCD, distance MCD, and energy captured ratio across different SNR ($N_{\text{MPC}} = 2$).}
  \label{fig:5}
\end{figure*}




\begin{figure}[htbp]
  \centering
	\includegraphics[width=\columnwidth]{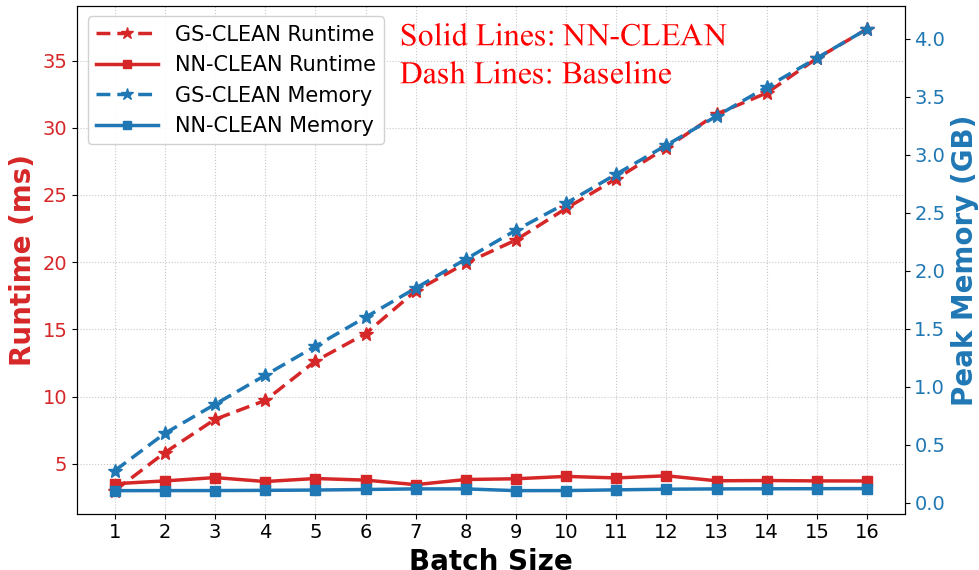}
  \caption{Runtime and Memory Comparison Between NN-CLEAN and GS-CLEAN}
  \label{fig:6}
\end{figure}

\subsection{Results and Analysis}

The proposed NN-CLEAN framework is evaluated via Monte Carlo simulations, comprising $10,000$ independent channel realizations per test configuration. To rigorously assess generalization against real-world propagation complexities, the testing data is synthesized entirely \textbf{off-grid}. Drawing the ground-truth parameters ($\phi_R, \phi_T, d$) from continuous uniform distributions---rather than discrete training bins---intentionally introduces realistic grid mismatch and spectral leakage to challenge the framework's robustness.

Under these stringent conditions, Fig.~\ref{fig:4} evaluates the multipath resolvability of NN-CLEAN against four established baselines: traditional GS-CLEAN \cite{seun_clean}, 3D-MUSIC \cite{3d_music}, 3D-ESPRIT \cite{3d_esprit}, and a standalone NN one-shot approach. Performance is benchmarked across varying MPC densities ($N_{\text{MPC}} \in \{1, 2, 3, 4\}$) at an SNR of 5~dB. Due to the spatial subspace limitations inherent to the simulated antenna configuration, evaluations for the subspace methods are strictly capped at $N_{\text{MPC}}=3$. Estimation accuracy is defined as the ratio of correctly identified MPCs to the total number of simulated paths. Specifically, an extracted MPC is considered a correct match if its predicted AoA, AoD, and distance simultaneously fall within a $\pm 2$ grid index tolerance of the ground truth. To explicitly isolate the efficacy of the proposed iterative framework, the one-shot baseline utilizes the identical NN architecture as NN-CLEAN but is trained to regress all $N_{\text{MPC}}$ components simultaneously in a single forward pass, assuming ideal heuristic pairing.

As shown in Fig.~\ref{fig:4}, the empirical results demonstrate that NN-CLEAN achieves peak estimation accuracy in sparse multipath scenarios ($N_{\text{MPC}} \in \{1, 2\}$), successfully bypassing the severe side-lobe artifacts that frequently limit traditional subspace and grid-search algorithms. As multipath density increases, both CLEAN-based approaches exhibit a linear decay in accuracy. Although GS-CLEAN degrades at a slightly slower rate---eventually surpassing NN-CLEAN in dense environments ($N_{\text{MPC}} \in \{3, 4\}$) due to cumulative error propagation within the NN's sequential extraction loop---the proposed framework maintains a substantial performance margin over the subspace methods. Crucially, NN-CLEAN consistently outperforms the standalone one-shot baseline across all tested densities. The severe degradation of the one-shot model explicitly highlights that embedding the NN within the iterative CLEAN loop is vital for mitigating accuracy loss as $N_{\text{MPC}}$ grows. Ultimately, because the dominant 1--2 paths capture the overwhelming majority of channel energy in practical deployments, NN-CLEAN's superior resolvability in these primary scenarios establishes it as a highly favorable framework for real-time systems.

Beyond resolving the existence of MPCs, it is critical to evaluate the absolute parameter precision of the identified paths. Fig.~\ref{fig:5} evaluates the estimation precision for a $N_{\text{MPC}}=2$ scenario across a sweep of $\text{SNR} \in [-5, 10]$~dB. Performance is explicitly quantified via three metrics: the Cumulative Distribution Function (CDF) of the Root Mean Square Error (RMSE) for the angular Multipath Component Distance (MCD) \cite{MCD_metric1} ($\eta_{\text{ang}}$), the distance MCD RMSE ($\eta_{d}$), and the overall energy captured ratio ($\epsilon$).

To prevent angular wrap-around discontinuities, the angular MCD between a ground-truth path and its estimate is formulated as:
\begin{equation}
\eta_{\text{ang}} = \frac{1}{2} \sqrt{ \left| e^{j\phi_{R}} - e^{j\hat{\phi}_{R}} \right|^2 + \left| e^{j\phi_{T}} - e^{j\hat{\phi}_{T}} \right|^2 } .
\end{equation}
Correspondingly, the normalized distance MCD is defined as:

\begin{equation}
\eta_{d} = \sqrt{ \left( \frac{|d - \hat{d}|}{\Delta d_{\text{max}}} \cdot \frac{d_{\text{std}}}{\Delta d_{\text{max}}} \right)^2 } .
\end{equation}
The overall angular and distance RMSEs are then computed from these individual path errors across all successfully matched components. Finally, the fidelity of the overall channel reconstruction is measured by the energy captured ratio $\epsilon = \left\| \sum_{i=1}^{N_{\text{MPC}}} \hat{\boldsymbol{H}}_i \right\|^2 \big/ \|\boldsymbol{H}\|^2$, where $\hat{\boldsymbol{H}}_i$ represents the channel matrix reconstructed from the $i$-th estimated MPC.

As depicted in Fig.~\ref{fig:5}, the empirical CDFs for the angular MCD (Fig.~\ref{fig:5}a), distance MCD (Fig.~\ref{fig:5}b), and energy captured ratio (Fig.~\ref{fig:5}c) demonstrate a shared trend: estimation fidelity predictably degrades as the SNR decreases. Specifically, the absolute error scales remain exceptionally small for both algorithms. Although the exhaustive GS-CLEAN method yields strictly superior precision across all evaluated metrics, the performance gap is remarkably narrow. The curves for NN-CLEAN closely align with the grid-search baseline, maintaining nearly identical distributions despite the tight scale of the plots. This confirms that while the NN sacrifices a marginal degree of precision due to the discretization of its output layer, it still provides highly accurate parameter estimations and reliable signal reconstruction.

Ultimately, this marginal precision trade-off is justified by a massive reduction in computational complexity. To isolate the fundamental per-iteration processing cost, Fig.~6 evaluates execution runtime and memory consumption for $N_{\text{MPC}}=1$ across a sweep of batch sizes. Both NN-CLEAN and GS-CLEAN are GPU-parallelized and evaluated on an NVIDIA GeForce RTX 4070 GPU with 12~GB of memory, with peak GPU memory recorded using TensorFlow statistics after warm-up. Here, the batch size denotes the number of independent channel realizations processed concurrently by a centralized processor, such as observations from multiple users or vehicles in an ISAC or multi-user MIMO system. The results reveal a stark divergence in scalability: at a minimal workload ($\text{Batch}=1$), both algorithms exhibit comparable execution times due to the fixed overhead of NN kernel initialization. However, even at this baseline, NN-CLEAN demonstrates superior architectural efficiency, consuming approximately half the memory of GS-CLEAN. As the batch size increases, the computational burden of GS-CLEAN escalates linearly, driven by the repetitive requirement to allocate high-resolution cost volumes and perform exhaustive 3D searches for every channel realization. In contrast, NN-CLEAN leverages the massive parallelization capabilities of the GPU to process multiple realizations in a single forward pass, maintaining a near-flat trend for both runtime and memory footprint. This significant amortized efficiency establishes NN-CLEAN as a highly scalable solution for high-throughput, real-time channel estimation in MIMO systems.

\section{Conclusion}
In this paper, we introduced NN-CLEAN, a novel hybrid framework that resolves the critical computational bottlenecks of the traditional CLEAN algorithm. By embedding a multi-head residual neural network directly into the iterative CLEAN extraction loop, the proposed method replaces exhaustive multi-dimensional grid searches with highly efficient, parallelizable inference. Evaluated under stringent off-grid conditions across varying multipath component ($N_{\text{MPC}}$) counts and SNR regimes, NN-CLEAN demonstrated superior resolvability compared to traditional subspace methods and standalone one-shot learning models. While the discretization of the network's output layer results in a marginal sacrifice in absolute precision compared to the exhaustive GS-CLEAN baseline, NN-CLEAN maintains exceptionally tight alignment in both parameter accuracy and signal reconstruction fidelity. Most importantly, this minor precision trade-off yields profound computational benefits. Unlike traditional methods whose processing and memory demands scale linearly with batch size, NN-CLEAN leverages GPU parallelization to achieve near-constant runtime and memory footprint. Ultimately, by anchoring the iterative subtraction of predicted MPCs to exact wave propagation physics, NN-CLEAN provides a highly robust, scalable, and physically grounded solution for real-time channel estimation in next-generation MIMO and ISAC systems.

\bibliographystyle{IEEEtran}
\bibliography{refs}

\end{document}